\documentclass{article}

\usepackage[preprint]{corl_2023} 

\usepackage{microtype}
\usepackage{graphicx}
\usepackage{tabularx}
\usepackage{wrapfig}
\usepackage{subfigure}
\usepackage{booktabs} 

\usepackage{hyperref}

\usepackage{amsmath}
\usepackage{amssymb}
\usepackage{mathtools}
\usepackage{amsthm}
\usepackage{dsfont}

\usepackage[capitalize,noabbrev]{cleveref}

\usepackage{algorithmicx}
\usepackage{algpseudocode}

\usepackage{bbm} 
\theoremstyle{plain}
\newtheorem{theorem}{Theorem}[section]

\newtheorem{lemma}[theorem]{Lemma}

\theoremstyle{definition}

\theoremstyle{remark}
\newtheorem{remark}[theorem]{Remark}

\usepackage[disable,textsize=tiny]{todonotes}

\usepackage{enumitem}  

\algnewcommand{\LineComment}[1]{\State \(\triangleright\) #1}
\algnewcommand{\LeftComment}[1]{\Statex \(\triangleright\) #1}

\newcommand{\U}{\mathcal{U}}
\newcommand{\X}{\mathcal{X}}
\newcommand{\Xsafe}{\mathcal{X}_{\mathrm{safe}}}
\newcommand{\Xunsafe}{\mathcal{X}_{\mathrm{unsafe}}}
\newcommand{\set}[1]{\{#1\}}
\newcommand{\R}{\mathbb{R}}

\title{Value Functions as Control Barrier Functions: Verification of Safe Policies using Control Theory}

\author{
  Daniel C.H.~Tan\\
  Department of Computer Science\\
  University College London\\
  United Kingdom\\
  \texttt{daniel.tan.22@ucl.ac.uk} \\
  \And
  Fernando Acero\\
  Department of Computer Science\\
  University College London\\
  United Kingdom\\
  \texttt{fernando.acero@ucl.ac.uk}\\
  \And
  Robert McCarthy\\
  Department of Computer Science \\ 
  University College London \\
  United Kingdom \\
  \texttt{robert.mccarthy.22@ucl.ac.uk} \\
  \And 
  Dimitrios Kanoulas \\
  Department of Computer Science \\ 
  University College London \\ 
  United Kingdom \\
  \texttt{d.kanoulas@ucl.ac.uk} \\
  \And 
  Zhibin Li\\
  Department of Computer Science\\
  University College London\\
  United Kingdom\\
  \texttt{alex.li@ucl.ac.uk}
}

\begin{document}
\maketitle

\vspace{-5mm}
\begin{abstract}
Guaranteeing safe behaviour of reinforcement learning (RL) policies poses significant challenges for safety-critical applications, despite RL's generality and scalability. To address this, we propose a new approach to apply verification methods from control theory to learned value functions. By analyzing task structures for safety preservation, we formalize original theorems that establish links between value functions and control barrier functions. Further, we propose novel metrics for verifying value functions in safe control tasks and practical implementation details to improve learning. Our work presents a novel method for certificate learning, which unlocks a diversity of verification techniques from control theory for RL policies, and marks a significant step towards a formal framework for the general, scalable, and verifiable design of RL-based control systems.
\end{abstract}

\section{Introduction}

Deep reinforcement learning (RL)~\cite{Sutton2018} is a powerful and scalable tool for solving control problems, such as Atari games~\cite{Mnih2013}, robotic control~\cite{Lillicrap2015}, and protein folding~\cite{Jumper2021}. However, because of their black-box nature, it is difficult to determine the behaviour of neural networks. In extreme cases, out-of-distribution or adversarially constructed inputs~\cite{Goodfellow2014} can catastrophically degrade network performance. In the control context, this can lead to highly unsafe behaviour; it is thus risky to deploy such controllers in safety-critical applications, such as autonomous vehicles or human-robot interaction, as well as future applications for general-purpose robots. 

The problem of safe control has been extensively studied in safe reinforcement learning, through the lens of constrained Markov Decision Processes~\cite{Altman1999}. Such methods implicitly assume that there are known constraints which are sufficient to guarantee safety. In contrast, our work assumes no prior knowledge of safe dynamics and aims to learn a constraint (in the form of a barrier function) to guarantee safety. This enables our approach to handle applications where safety cannot be easily expressed analytically, such as avoiding dynamic obstacles from raw pixel input \cite{Dawson2022perception}. 

On the other hand, there exists rich literature in control theory on proving properties of dynamical systems using \textit{certificate functions}. The most well-known are Lyapunov functions, which prove the stability of dynamical systems around a fixed point~\cite{isidori1985nonlinear}. Traditionally, it is difficult to design certificate functions for complex systems of interest. We discuss recent learning-based methods in Section~\ref{section:related_work}. Other prior work combines classical and RL-based control methods by learning high-level policies over programmatic low-level controllers~\cite{Margolis2022}, which could be designed to respect safety constraints. 

However, designing effective and safe low-level controllers is still difficult and time-consuming. In both cases, the difficulty of manual design limits scalability to arbitrary tasks. Drawing inspiration from control theory, we aim to design a learning-based control method that benefits from the verifiability of certificate functions without sacrificing the generality and flexibility of reinforcement learning.
    
\begin{figure}[t]
    \vspace{-8mm}
    \centering
    \includegraphics[width=.9\linewidth]{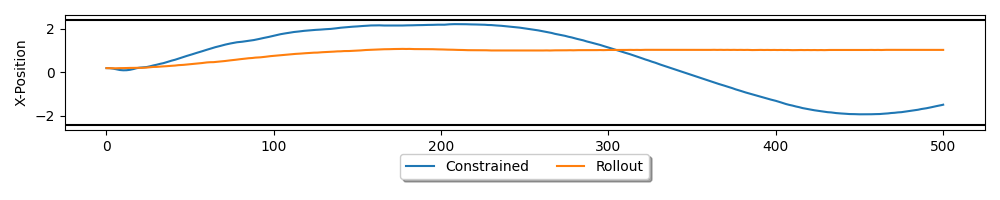}
    \vspace{-5mm}
    \includegraphics[width=.9\linewidth]{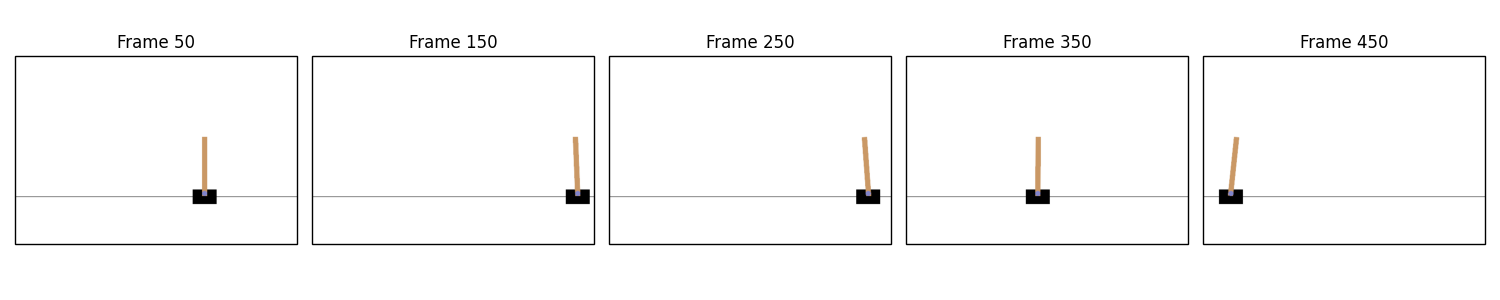}
    \caption{Top: Control barrier function (CBF) constrained exploration (blue) allows reaching both extremes of the CartPole safe state space (demarcated in black). Rolling out reward-optimal policy fails to do so (orange). Bottom: The CBF-constrained trajectory is visualized.}
    \label{fig:cbf_constrain_viz}
    \vspace{-3mm}
\end{figure}

Our contributions are twofold. Firstly, we propose a \textbf{reinforcement learning method} for synthesizing control barrier certificates. Under mild assumptions on task structure, we prove a strong connection between barrier functions and value functions. We implement and ablate principled design choices for learning good barrier functions. Secondly, we propose and empirically validate novel \textbf{metrics} to evaluate the quality of learned barrier functions. We demonstrate that these metrics capture important structure not reflected in standard RL metrics. Control barrier certificates verified by these metrics successfully allow safety-constrained exploration of a large fraction of the safe state space, as shown in Figure~\ref{fig:cbf_constrain_viz}.

Concretely, our method involves considering a safety-preserving task, where the reward function is given by $r = 0$ in safety-violating states and $r=1$ otherwise. We show that the value function $V$ satisfies properties to be a control barrier function and we derive a threshold for predicting safety. We then learn $V$ by using standard RL techniques, propose new metrics to verify learned $V$ as a control barrier function, and finally demonstrate that our metrics capture the safety-preserving capacity of $V$. By connecting value functions to certificate functions, our work presents a novel perspective on learning certificate functions, which offers a new approach for applying the wealth of verification strategies in control theory to reinforcement learning.

\section{Preliminaries}
In this work, we consider deterministic Markov Decision Processes (MDPs) and reinforcement learning (RL), with states $x \in \X$ and actions $u \in \mathcal{U}$. Further exposition is provided in Appendix~\ref{appendix:preliminaries}.

\subsection{Indefinitely Safe Control}
\begin{wrapfigure}{r}{0.5\textwidth}
    \centering
    \includegraphics[width=\linewidth]{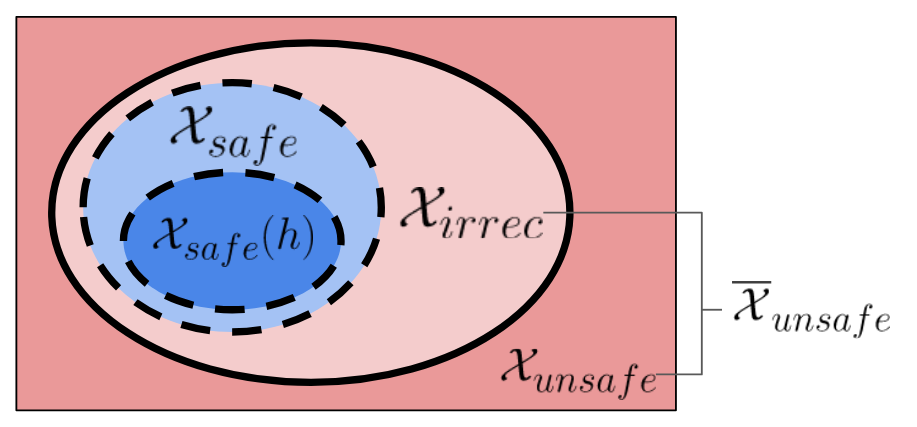}
    \caption{Partitioning of $\X$. $\Xunsafe$ is specified \textit{a-priori}; $\Xsafe, \X_{\mathrm{irrec}}$ are then uniquely defined by $(f, \Xunsafe)$. For any valid CBF $h$, we have $\Xsafe(h) \subseteq \Xsafe$. }
    \label{fig:x_partition}
\end{wrapfigure}

We consider augmenting an MDP with a set of \textit{safety violations} $\Xunsafe$, unsafe states specified by the practitioner. This partitions the state space $\X$ into three subsets $\Xunsafe, \Xsafe, \X_{\mathrm{irrec}}$, illustrated in Figure~\ref{fig:x_partition}. $\Xsafe$ consists of \textit{indefinitely safe} states; i.e., there exists a controller $\pi: \X \to \U$ such that $\Xsafe$ is \textit{forward-invariant} under closed-loop dynamics $f_\pi(x) = f(x, \pi(x))$. $\X_{\mathrm{irrec}}$ consists of \textit{irrecoverable} states. For example, a car travelling at high velocity on a low-friction surface may inevitably collide with an imminent obstacle despite applying maximum braking effort. We define $\overline{\X}_{\mathrm{unsafe}} = \X_{\mathrm{unsafe}} \cup \X_{\mathrm{irrec}}$. 

\textbf{Finite irrecoverability}. In general, due to $\X_{\mathrm{irrec}}$, safe control requires perfect knowledge of dynamics for arbitrarily long horizons, which can be intractable; hence we assume stronger conditions on $\X_{\mathrm{irrec}}$. We say $x$ is $k$-irrecoverable if it is guaranteed to enter $\Xunsafe$ within $k \in \mathbb{N}$ timesteps regardless of control. For $x \in \X_{\mathrm{irrec}}$, let $k_x$ be the minimum integer such that $x$ is $k_x$-irrecoverable. We will assume $\set{k_x: x \in \X_{irrec}}$ is upper bounded by a constant $H < \infty$. Finite irrecoverability has been studied in previous work \cite{Thomas2022}, and is expected to be satisfied for reasonably well-actuated dynamics $f$ and well-behaved choices of $\Xunsafe$. 

\subsection{Control Barrier Functions}
Control barrier functions (CBFs) are a useful tool for solving safe control problems. A CBF $h: \X \to \R$ can be thought of as a classifier that classifies safe and unsafe states according to its level set $h(x) = 0$. The set $\set{x: h(x) \geq 0}$ defines a safe set $\Xsafe(h)$. Loosely speaking, larger values of $h(x)$ correspond to `safer' states. Formally, given $(M, \Xunsafe)$, and $\alpha \in (0,1]$, we say that $h: \X \to \R$ is a (discrete-time) CBF \textit{against} $\Xunsafe$ if it satisfies: 
\begin{equation}\label{def:cbf}
   \begin{aligned}
      &(i) \quad \forall x \in \Xunsafe, \quad h(x) < 0   \\
      &(ii) \quad \forall x: h(x) \geq 0, \quad \sup_u \set{h(f(x, u))} \geq (1-\alpha) h(x)
   \end{aligned}
\end{equation}
We note the following properties, proved in the appendix:

\begin{lemma}\label{thm:h_properties} By condition (\ref{def:cbf})(i), $\Xsafe(h) \cap \Xunsafe = \emptyset$. By condition (\ref{def:cbf})(ii), there exists a policy $\pi$ such that $\Xsafe(h)$ is \textit{forward-invariant} under $f_\pi$; if $x \in \Xsafe(h)$, then  $f(x, \pi(x)) \in \Xsafe(h)$.
\end{lemma}

A CBF $h$ is useful for safe control because it eliminates the need to reason about dynamics over long horizons. Instead, we only need to check a one-step bound in condition (\ref{def:cbf})(ii) to guarantee safety indefinitely. One edge case occurs when $\Xsafe(h) = \emptyset$; we call such CBFs \textit{trivial}. Subsequently we assume that we can always find nontrivial CBFs against $\Xunsafe$ (if not, this indicates that $\Xunsafe$ is `too large' and we should reconsider the choice of $\Xunsafe$).

\textbf{Transforms of CBFs}. Lastly, we note that certain classes of transformations preserve the control barrier function property, formalized as follows: 
\begin{lemma}\label{thm:h_transform}
Let $h: \X \to \R$ be a CBF. Let $w: \R \to \R$ such that $\mathrm{Im}(h) \subseteq \mathrm{Dom}(w)$. Suppose there exists $C \geq 0$ such that for all $x, y \in \mathrm{Dom}(w)$, we have: 
\begin{equation}\label{eqn:h_transform}
   \begin{aligned}
      &(i) \quad w(x) \geq Cx   \\
      &(ii) \quad w(x) - w(y) \geq C(x-y) \\
      &(iii) \quad \set{x: w(x) \geq 0} = \set{x: x \geq 0}. 
   \end{aligned}
\end{equation}
Then $\tilde{h} = w \circ h$ is also a CBF. We will say that such $w$ are \textit{CBF-preserving} transforms.
\end{lemma}

The proof is straightforward and given in the appendix. 

\section{Learning of Control Barrier Functions}

This section presents the main results connecting value functions to control barrier functions, and then proposes a principled and practical algorithm for learning control barrier functions. Detailed proofs can be found in Appendix~\ref{appendix:proofs}. 

\subsection{Safety Preserving Task Structure}\label{theory:safety_preservation}

In the safety-preserving task framework, we assume a reward structure of: $r(x, u, x') = 0$ when $x' \in \Xunsafe$; otherwise, $r(x, u, x') = 1$. We also assume \textit{early termination}, where the episode terminates immediately when $x' \in \Xunsafe$. Within this task structure, we analyze the optimal value function $V^*$ under the partition illustrated in Figure \ref{fig:x_partition}, which consists of:
\begin{itemize}[leftmargin=*]
    \item $x \in \Xunsafe$. Since the episode terminates immediately, we trivially have $V(x) = 0$.   
    \item $x \in \Xsafe$. In this case, we know there exists a policy which preserves safety indefinitely, hence we have $V(x) = \sum_{j=0}^\infty \gamma^j(1) = \frac{1}{1-\gamma}$. 
    \item $x \in \X_{\mathrm{irrec}}$. Let $x$ be $k$-irrecoverable. Then $V^*(x) = \sum_{j=0}^{k-1} \gamma^j = \frac{1-\gamma^k}{1-\gamma}$. 
\end{itemize}

We make two remarks from this analysis. Firstly, $V^*$ is \textit{bounded} ; we have $V^*(x) \in [0, \frac{1}{1-\gamma}]$. Secondly, the range of $V^*$ is \textit{partitioned} by $\Xsafe, \overline{\X}_{\mathrm{unsafe}}$:
 \begin{equation}\label{eqn:v_partitioned}
     \sup_{\overline{\X}_{\mathrm{unsafe}}}\set{V^*(x)} = \frac{1 - \gamma^H}{1-\gamma} < \frac{1}{1-\gamma} = \inf_{\Xsafe} \set{V^*(x)}
 \end{equation}
These two observations motivate us to propose CBFs of the form $h = V^* - R$, formalized below. 
 
\begin{theorem}\label{thm:v_as_cbf}
    Let $M$ be an MDP and suppose (a) early termination is employed with termination condition $\Xunsafe$, (b) $r$ has safety-preserving reward structure, and (c) there exists an upper bound $H$ on irrecoverability. Then for any $R \in (\frac{1 - \gamma^H}{1-\gamma}, \frac{1}{1-\gamma}]$, we have that $h = V^* - R$ is a control barrier function against $\Xunsafe$.
\end{theorem}

In practice, we do not have access to $V^*$; we only have access to learned functions $V \approx V^*$. Nonetheless, so long as $V$ is `not too far' from $V^*$, we can use $h = V(x) - R$ as a barrier function. 
\begin{theorem}\label{thm:v_learned_as_cbf}
    Let $M$ be an MDP and let the assumptions (a) - (c) of Theorem \ref{thm:v_as_cbf} hold. Additionally, assume that $V$ satisfies (d) $\epsilon$-optimality; $\sup_{x \in X} |V(x) - V^*(x)| < \epsilon$, (e) $\epsilon < \frac{\gamma^H}{2(1-\gamma)}$. Then for $\alpha \in [\frac{2\epsilon}{\frac{1}{1-\gamma} + \epsilon - R}, 1]$ and any $R \in (\frac{1-\gamma^H}{1-\gamma} + \epsilon, \frac{1}{1-\gamma} - \epsilon]$, we have that $h = V - R$ is a control barrier function against $\Xunsafe$.
\end{theorem}

We find that the bound on $\epsilon$ is very permissive. To illustrate how loose the bound is, let $H = 10$ \cite{Thomas2022} and $\gamma = 0.99$. Then $\epsilon \leq \frac{1-\gamma^H}{2(1-\gamma)} \approx 47$ suffices, inducing a corresponding $R = \frac{1}{1-\gamma} - \epsilon \approx 53$ and $\alpha = 0.96$. For smaller $\epsilon$, a wider range of values of $R$ will be valid. In our experiments we find that $R = \frac{1}{2(1-\gamma)} = 50$ and $\alpha = 0.1$ work well empirically. Note that in our approach, we do not need to explicitly set $H$; rather, it is defined implicitly by $R$.  

\subsection{Reinforcement Learning Framework}

We train a Deep Q-Network~\cite{Mnih2013} for $2 \times 10^6$ timesteps on the CartPole environment in OpenAI Gym \cite{brockman2016openaigym}. A detailed description is provided in Appendix \ref{appendix:cartpole}. The network parametrizes a Q-function; the corresponding value function is $V(x) = \sup_{u \in \U} Q(x,u)$. The network is trained via standard temporal-difference learning \cite{Sutton2018} to minimize the TD error: $\mathcal{L}_{TD} = \mathbb{E}_{(x, u, x') \sim f_\pi} \|r(x,u) + \gamma V(x') - V(x)\|^2$. Our baseline uses the implementation in CleanRL \cite{Huang2022cleanRL}. Training results are visualized in Appendix \ref{appendix:figures}.

\textbf{Implementation details.} In theory, training a sufficiently expressive $V$ for sufficiently long on the TD objective results in $V$ converging uniformly to $V^*$. In practice, we find that training a vanilla DQN is insufficient; certain additional implementation details are required to obtain high-quality barrier functions. Below, we describe and motivate these design choices. 

\textbf{Bounded value}. Recall from Section~\ref{theory:safety_preservation} that $V^*$ is bounded; this motivates us to consider a parametrization of the form $V(x) = g(\sigma(\phi(x)))$ where $\phi$ is a neural network, $\sigma(x) = \frac{1}{1 + \exp(-x)}$ is the sigmoid function and $g(x) = \frac{x}{1-\gamma}$ is a linear mapping that allows $V$ to have the correct range.  We hypothesize that this aids learning by stabilizing the learning signal on the network weights, by essentially converting the two-sided regression loss into a one-sided classification loss. We denote architectures with bounded value by \texttt{SIGMOID}, and those without by \texttt{MLP}. 

\textbf{Supervision of $V$}. Recall that we analytically know $V^*(x) = 0$ for $x \in \Xunsafe$. This motivates us to introduce a supervised loss $\mathcal{L}_{\mathrm{unsafe}} = \mathbb{E}_{x \sim \Xunsafe} \| V(x) \|$. Since we can specify $\Xunsafe$, this loss can be approximated by sampling $\Xunsafe$ (e.g. by rejection sampling). We hypothesize that the supervised loss provides a valuable auxiliary learning signal that complements the standard TD objective. Because it is undesirable to enter unsafe states, we expect such states to be sparsely sampled. Furthermore, due to early termination, it may be outright impossible to reach most of $\Xunsafe$ (beyond a thin boundary). Hence, the supervised loss over $\Xunsafe$ provides a learning signal exactly in the regions where the TD objective does not, and vice versa. We indicate models trained with supervision by \texttt{\{SIGMOID,MLP\}-SUP}.

\textbf{Exploration}. We implement stronger exploration by modifying the initial state distribution to be more diverse. Because the TD objective only acts on states experienced during rollout, improved exploration provides a learning signal to $V$ over a larger region of $\X$. Exploration through diverse reset initialization is enabled by default; to evaluate its impact, we perform an experiment using the original state distribution, denoted by \texttt{NOEXP}. 

\begin{table}
\centering
\vspace{-8mm}
\begin{center}
\begin{small}
\begin{sc}
\begin{tabular}{l|ccc|cc}
\toprule
Experiment & MLP & SIGMOID & MLP-SUP & SIGMOID-SUP & NOEXP \\
\midrule
Bounded & No & Yes & No & Yes & Yes \\
Supervised & No & No & Yes & Yes & Yes \\
Exploration & Yes & Yes & Yes & Yes & No \\
\midrule
$\pi^*$ return & $493 \pm 14.8$ & $\textbf{500} \pm 0$ & $465 \pm 48.0$ & $\textbf{500} \pm 0.0$ & $\textbf{500} \pm 0.0$ \\
TD error & $2.43 \pm 0.71$ & $2.09 \pm 0.27$ & $0.958 \pm 0.14$ & $0.746 \pm 0.057$ & $\textbf{0.607} \pm 0.053$ \\
$m_{valid}(h)$ & $0.476 \pm 0.140$ & $0.752 \pm 0.130$ & $0.603 \pm 0.046$ & $0.991 \pm 0.002$ & $\textbf{0.993} \pm 0.002$ \\
$m_{cov}(h)$ & $\textbf{0.767} \pm 0.146$ & $0.477 \pm 0.141$ & $0.595 \pm 0.048$ & $0.106 \pm 0.010$ & $0.063 \pm 0.013$ \\
$\pi_h$ return & $9.36 \pm 0.16$ & $21.3 \pm 14.4$ & $21.3 \pm 11.2$ & $\textbf{163.5} \pm 54.7$ & $114.6 \pm 85.1$ \\
\bottomrule
\end{tabular}
\end{sc}
\end{small}
\end{center}
\caption{Description and final metrics for $5$ seeds of $5$ settings. On all metrics except coverage, enabling both bounded parametrization and supervision outperformed all ablations. The lower coverage can be explained by the trade-off between $m_{valid}$ and $m_{cov}$.} 
\label{table:experimental_settings}
\end{table}

\begin{figure*}
\vspace{-5mm}
    \centering
    \includegraphics[width=\linewidth]{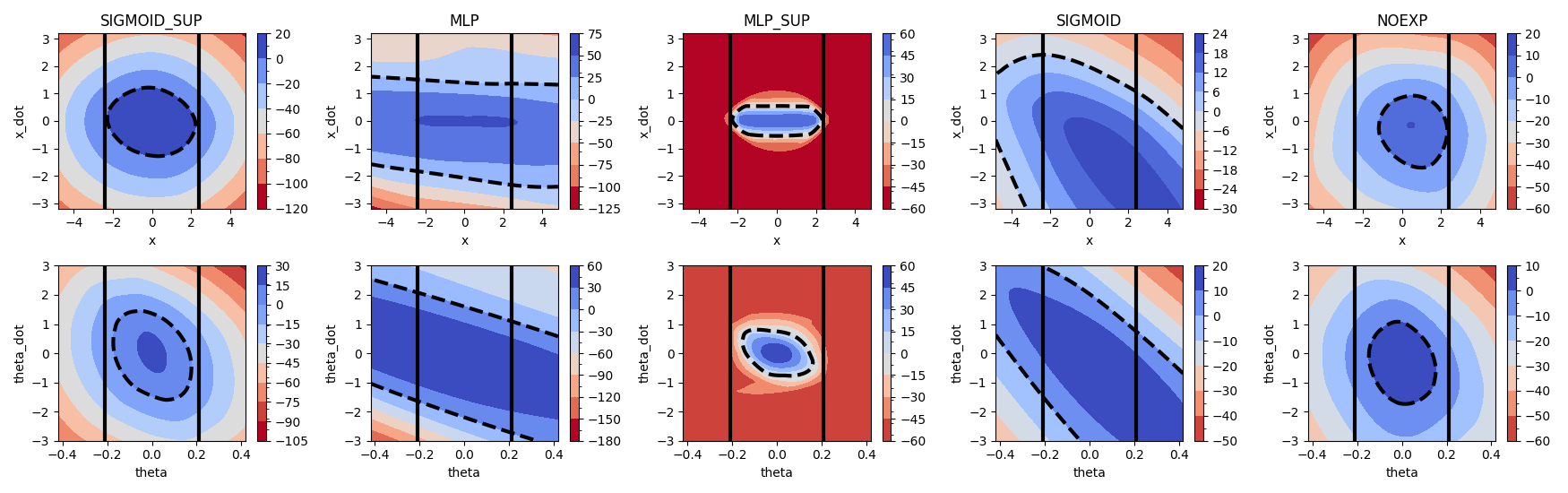}
    \caption{Phase diagram of learned control barrier functions for various experimental settings. In all cases, we use $R = {1}/{(2(1-\gamma))} = 50$. Top: $h(x, \dot{x}, 0, 0)$ for varying $x, \dot{x}$. Bottom: $h(0, 0, \theta, \dot{\theta})$ for varying $\theta, \dot{\theta}$. Solid lines demarcate $\Xunsafe$. Dashed lines indicate $\Xsafe(h) = \set{x: h(x) \geq 0}$. Overall, the \texttt{SIGMOID-SUP} model is qualitatively best.}
    \label{fig:viz_cbf}
\vspace{-5mm}
\end{figure*}

Recalling Lemma \ref{thm:h_transform}, we define barrier functions of the form $h = w(V(x)-R)$ with $R = \frac{1}{2(1-\gamma)}$. In the case where unbounded value functions are used, we let $w$ be identity;  i.e. $h(x) = V(x)-R$. In the case where bounded value functions are used, we define $h(x) = \phi(x) = w(V(x)-R)$. The corresponding transform is $w(x) = \tilde{\sigma}^{-1} \circ g^{-1}$, with $\tilde{\sigma} = \sigma(x) - 0.5$. We assert that $w$ is CBF-preserving; a proof is given in the Appendix. 

\begin{remark}\label{remark:w_cbf_preserving} $w = \tilde{\sigma}^{-1} \circ g^{-1}$ is CBF-preserving. 
\end{remark}

\section{Verification of Learned CBFs}

After obtaining candidate barrier functions through the learning process, it is crucial to verify whether they meet the conditions in (\ref{def:cbf}). We investigate a total of $5$ experimental settings, ablating each design choice, summarized in Table \ref{table:experimental_settings}, and perform $5$ seeded runs of each setting. We visualize the learned barrier functions for each setting in Figure \ref{fig:viz_cbf}. Overall, the \texttt{SIGMOID-SUP} model is best. Supervision is essential to ensuring $\Xsafe(h) \cup \Xunsafe = \emptyset$. Exploration results in a larger $\Xsafe(h)$. We also remark that \texttt{SIGMOID} results in more even contours than \texttt{MLP}. 

Despite clear differences in CBFs between model variants, we note that standard metrics used in RL such as episode return and TD error fail to capture this discrepancy, as evidenced in Appendix \ref{appendix:figures} and Figure \ref{fig:train_hist_cbf}. Therefore, we further propose metrics that evaluate the quality of learned barrier functions. 

\subsection{Metrics on Control Barrier Functions}
\textbf{Validity.} Given $h$, we aim to quantify the extent to which it is valid across the state space, satisfying the conditions in (\ref{def:cbf}). Concretely, we will define a \textit{validity} metric $m_{\mathrm{valid}}$ to measure the quality of the learned CBF. Hence we rewrite (\ref{def:cbf}) as logical assertions $p_i$ and define associated predicates $\rho_i: \X \to \{0, 1\}$ indicating whether $p_i$ holds for $x \in \X$. 
\begin{itemize}[leftmargin=*]
    \item $p_1(x) := x \in \Xunsafe \implies h(x) < 0$. We define the associated predicate $\rho_1(h) = 1 - \mathds{1}\set{x \in \Xunsafe, \, h(x) \geq 0}$. 
    \item $p_2(x, \alpha) := \, h(x) \geq 0 \implies \sup_u \set{h(f(x,u))} \geq (1-\alpha)h(x)$. We define the associated predicate $\rho_2(h, \alpha)  =  1 - \mathds{1}\set{h(x) \geq 0, \, \sup_u \set{h(f(x,u))} < (1-\alpha)h(x)}$. 
\end{itemize}
We have defined $\rho_1, \rho_2$ such that $\mathbb{E}_{x \in \X}[\rho_1(h)(x)]$ (respectively $\rho_2(h, \alpha)$) measures the fraction of states where condition (\ref{def:cbf})(i) (respectively (\ref{def:cbf})(ii)) holds. Since we need both conditions to hold for $h$ to be a barrier function, it makes sense to define the metric $m_{\mathrm{valid}}(h) = \mathbb{E}_{x \in \X}[\rho_1(h)(x)\rho_2(h,\alpha)(x)]$. In all experiments, we use a value of $\alpha = 0.1$. 

\textbf{Coverage.} Given $h$, we would also like to measure the size of its safe set. A trivial barrier function (where $\Xsafe(h) = \emptyset$) is of no practical use even if it is valid everywhere. We measure this with the \textit{coverage} metric $m_{\mathrm{cov}}(h) = \mathbb{E}_{x \in \X}[\mathds{1}\set{h(x) \geq 0}]$, computed by sampling. In practice, we sample from a bounded subset $\X'$ which is assumed to contain $\Xsafe$. 

\begin{wrapfigure}{r}{0.5\textwidth}
    \vspace{-5mm}
    \centering
    \includegraphics[width=0.95\linewidth]{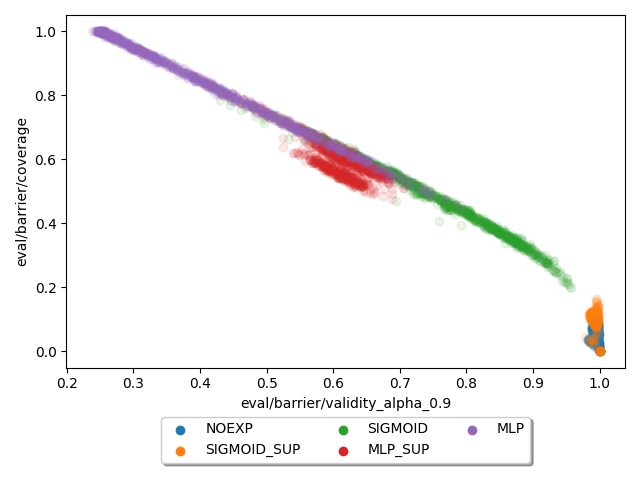}
    \caption{Validity and coverage throughout training history of different architectures. As validity increases, coverage tends to decrease. The best barrier functions have $m_{valid} = 1$, and $m_{cov}$ as high as possible subject to that.}
    \label{fig:tradeoff_validity_coverage}
    \vspace{-2mm}
\end{wrapfigure}

\textbf{Discussion.} Throughout the training history of different architectures, we observe a trade-off between validity and coverage, demonstrated in Figure \ref{fig:tradeoff_validity_coverage}. Validity refers to the extent to which a barrier function satisfies the specified conditions, while coverage measures the proportion of the state space on which the barrier function is applicable.
The goal is to find the best barrier functions that achieve a validity metric, $m_{\mathrm{valid}}$, equal to 1, indicating complete satisfaction of the conditions. Simultaneously, we aim to maximize the coverage, measured by the metric $m_{\mathrm{cov}}$, while still maintaining the high validity. We visualize training histories of $m_{\mathrm{cov}}, m_{\mathrm{valid}}$ in Figure \ref{fig:train_hist_validity_coverage} of Appendix \ref{appendix:figures}. The final results are also summarized in Table \ref{table:experimental_settings}. Empirically, bounded parametrization and supervision both aid in improving validity, whereas exploration aids in improving coverage. Thus, our experimental design choices are vindicated by evaluation on barrier metrics. More importantly, we note that standard RL metrics such as episode reward and TD error did not accurately distinguish between the learned networks in this regard. This demonstrates that our proposed barrier metrics provide a valuable and \textit{orthogonal} perspective for evaluating learned barrier functions. 

\subsection{Safety Constraints with Barrier Functions}

One common use of control barrier functions is to constrain a nominal policy $\pi_{nom}$ to respect safety constraints. While this naively requires one-step lookahead to calculate $h(f(x,u))$, we note that the Q-function allows us to perform \textit{implicit} one-step lookahead through the Bellman optimality condition $Q^*(x,u) = r(x,u) + \gamma V^*(f(x,u))$.  Thus, we define the safety-constrained policy:
\begin{equation}
    \pi_{h}(x) = 
    \begin{cases}
        \pi_{nom}(x) & \text{if } Q(x, \pi_{nom}(x)) \geq R \\
        \mathrm{argmax}_u Q(x,u) & \text{if } Q(x, \pi_{nom}(x)) < R
    \end{cases}
\end{equation}

\begin{figure}
    \vspace{-15mm}
    \centering
    \includegraphics[width=\linewidth]{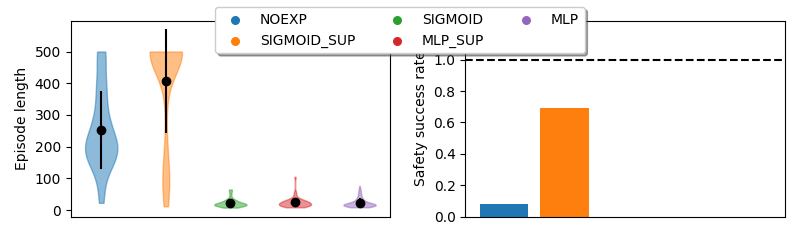}
    \caption{Safety-constrained exploration with learned CBFs. Left: Average safety-constrained episode length over $100$ rollouts. Right: Safety success rate, defined as fraction of episodes with no safety violations. \texttt{SIGMOID-SUP} is the best safety constraint. }
    \label{fig:eval_safety_constrain}
    \vspace{-5mm}
\end{figure}

We take $\pi_{nom}$ to be the uniform random policy and roll out $100$ episodes of $\pi_h$ for varying $h$. For each architecture, we evaluate (i) the safety-constrained episode length, and (ii) the safety success rate, defined as the fraction of episodes without safety violations. The results are summarized in Figure \ref{fig:eval_safety_constrain}. On the whole, the architectures \texttt{SIGMOID-SUP,NOEXP} with higher validity $m_{valid}$ serve as better safety constraints, justifying the use of $m_{valid}$ for model selection. However, we note that the best architecture failed to reach a safety success rate of $100\%$; we attribute this to the fact that $m_{valid}$ is not a rigorous measure of validity, but only provides statistical evidence of validity through sampling. 

\todo{(make a plot with several safety thresholds and demonstrate that increasing the threshold corresponds to more conservative policies.)}

\section{Additional Experiments on Robotics Environments}\label{section:additional_experiments}

Our case study on the CartPole environment demonstrates that the 0-1 safety reward induces an analytically tractable control barrier function with a safety threshold that is known \textit{a priori}. This has two important consequences: (i) the learned Q-value function can be used as a safety constraint and (ii) the CBF can be formally verified by checking control barrier function properties. We now present additional empirical results for (i) on 3 benchmark locomotion environments implemented in MuJoCo \cite{todorov2012mujoco}. Formal verification of control barrier functions in high-dimensional continuous-control systems is a promising direction that we leave open for future work. 

\subsection{Methodology}

We now describe important differences in methodology from our CartPole case study, related to scaling up. 

\textbf{Variational value function.} In DQN, the policy is implicitly parametrized as $\mathrm{argmax}_u Q(x,u)$. In an environment with a high-dimensional continuous control space, this becomes infeasible to evaluate; hence, it is typical to use an actor-critic architecture with a separate policy network $\pi(x)$. Correspondingly, our CBF is now approximated by a variational lower bound $V(x) \geq Q(x, \pi(x))$. Notably, this does not affect the justification for bounding or supervision; if $V$ is bounded, then $Q$ will still be bounded; furthermore, on the unsafe set, $V(x) = 0$ still implies $Q(x, \pi(x)) = 0$ (and indeed $Q(x,u) = 0$ for any $u$). 

\textbf{Offline RL}. Within our framework, the CBF is learned entirely offline with respect to the final safety-constrained policy. For this experiment, we therefore consider the fully-offline setting. Learning from offline data has the advantage that no additional safety violations must be encountered to learn the safety constraint. We leverage the datasets from D4RL \cite{fu2021d4rl}. Concretely, we train an actor-critic architecture with the TD3-BC algorithm \cite{fujimoto2021minimalist}. We relabel offline trajectories with the 0-1 safety-preserving reward defined above and evaluate the learned CBFs on the mean safety-constrained episode length under a random policy. Our suggested implementation details of bounding and supervision are used as before. 

\begin{figure}[h]
    \centering
    \includegraphics[width=0.32\textwidth]{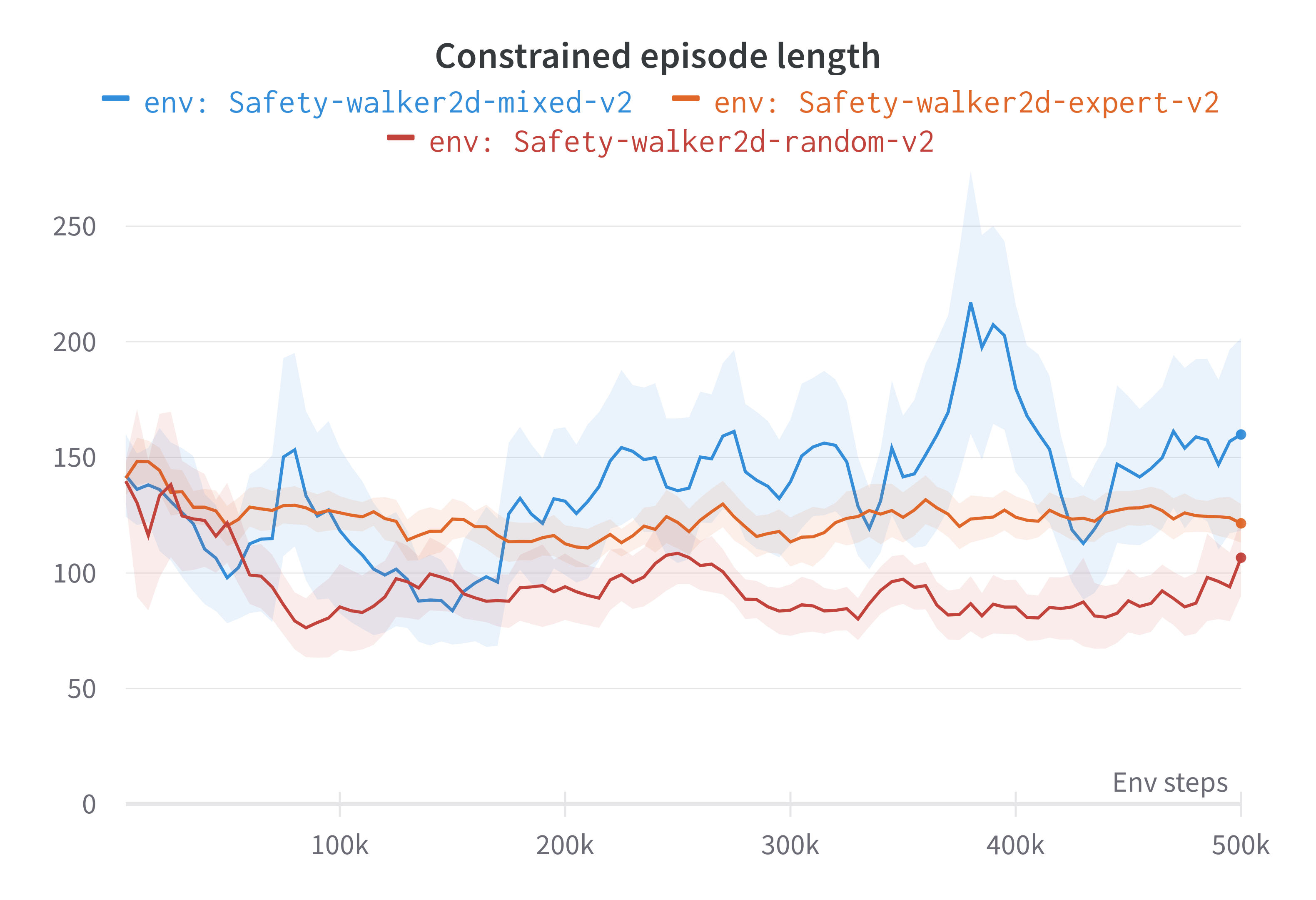}
    \includegraphics[width=0.32\textwidth]{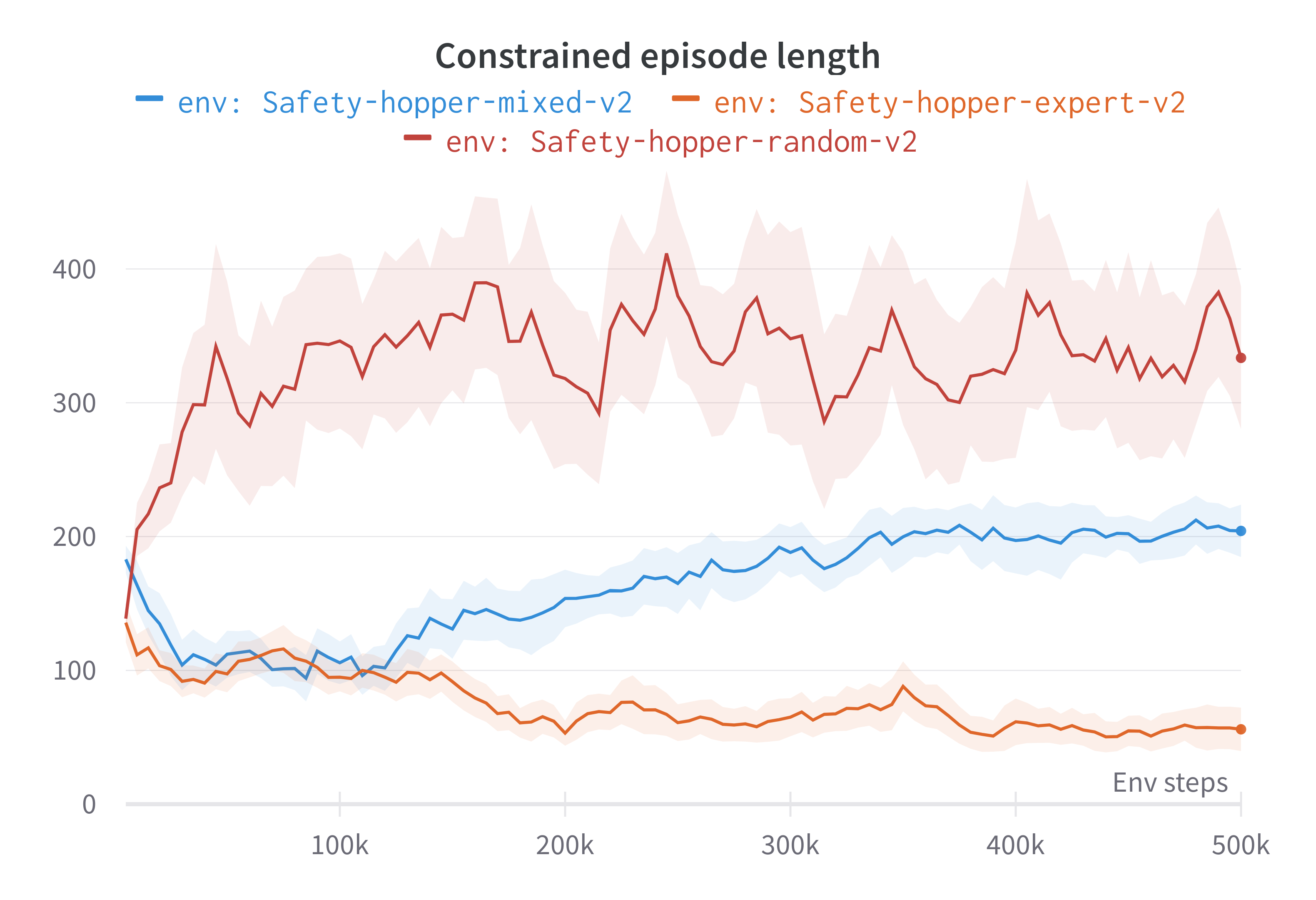}
    \includegraphics[width=0.32\textwidth]{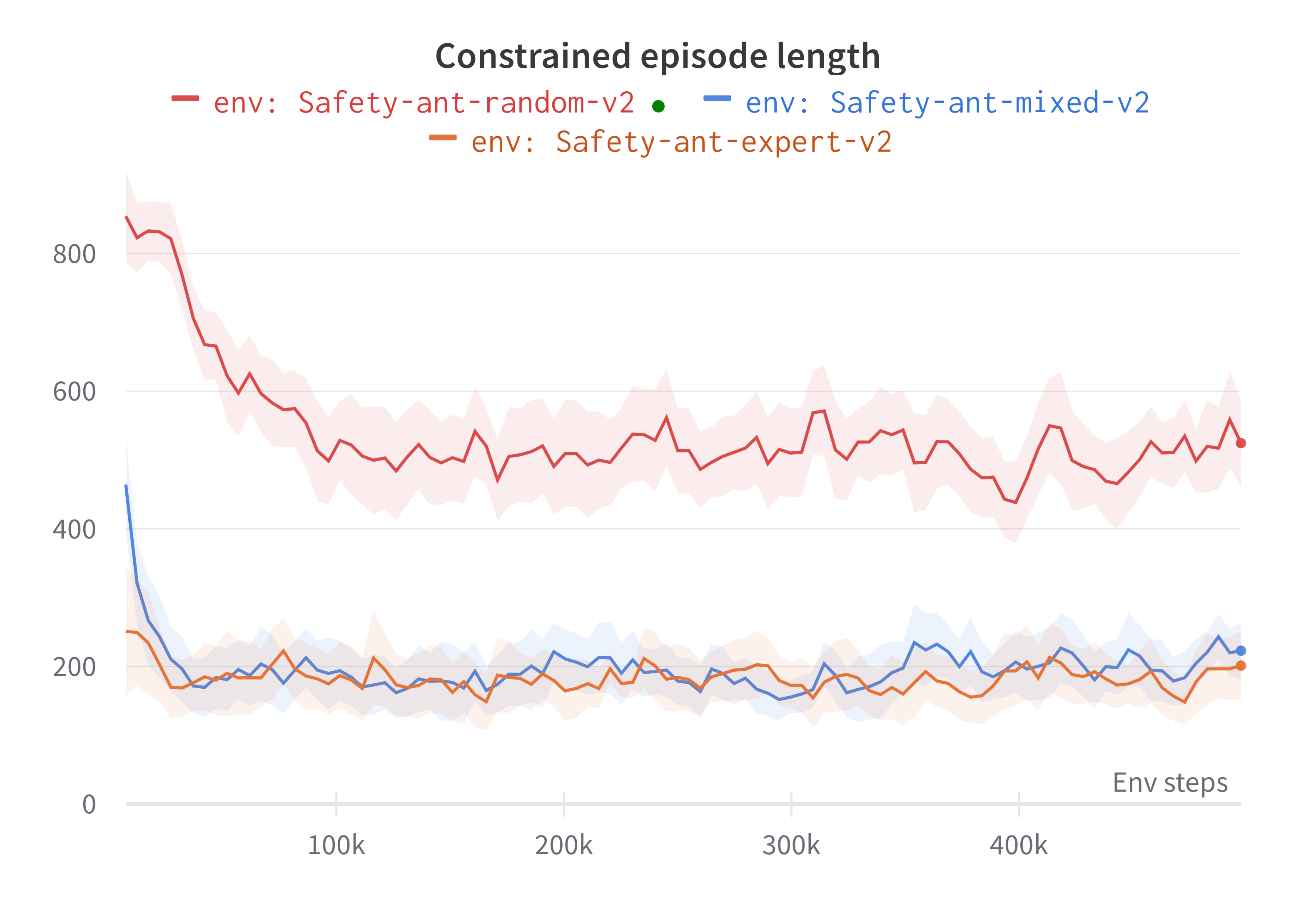}
    \caption{Training history of safety-constrained episode length, measured as the number of timesteps that the learned CBF preserves safety under a random policy, with standard error measured across 5 seeded experiment runs for Walker2d, Hopper, and Ant. The random dataset performs well across all environments and beats all other alternatives on 2 out of 3 environments. Unconstrained episode lengths are not shown here but typically do not exceed 30. }
    \label{fig:locomotion-dset-expt}
\end{figure}

\textbf{Choice of offline data.} We now describe considerations on the choice of dataset to use for offline RL. The main consideration is that CBF needs to robustly detect safety in the state distributions encountered by random actions. This is best achieved when the training data contains rollouts from a random policy. To study this effect, we test various choices of dataset: (i) \textit{random} data from a random policy, (ii) \textit{expert} data containing rollouts from a policy that achieves exert performance on the base locomotion task, and (iii) \textit{mixed} data containing a mixture of data of varying skill levels. The results are available in Figure \ref{fig:locomotion-dset-expt}. Empirically, we find that using only random data results in strong safety-constrained performance across a variety of environments, beating all other baselines in 2 out of 3 tested environments. 

One other consideration on the dataset involves the quality of policy behaviour. Since $Q(x, \pi(x))$ is a variational lower bound of $V(x)$, the approximation error is minimized when $\pi(x)$ maximizes $Q(x, \cdot)$, which occurs when $\pi$ is an expert safe policy. When $\pi$ is of lower quality, the typical values attained by the CBF are lower, resulting in a more conservative CBF. This has the benefit of being better at preserving safety but may be overly restrictive for exploration. We leave explicit investigation of this effect to future work. 

\section{Related work}\label{section:related_work}

\textbf{Safety analysis of RL.} Our work adds to a body of existing theoretical work on safety analysis for learning-based control. Safe MBPO \cite{Thomas2022} derived analytic safety penalties to guarantee safety within the model-based setting. Safe value functions, proposed in \cite{Massiani_2023}, provide a general reward framework under which reward-optimal policies are guaranteed to satisfy state constraints. Our work builds upon these results by (i) extending to the model-free setting; and (ii) providing a practical method of using learned critics as a safety constraint for downstream tasks. A separate line of work considers extensions to stochastic MDPs using the theory of almost Lyapunov functions \cite{huh2020safe}. We consider this to be a promising direction for future work within our framework. 

\textbf{Certificate-based RL.} Previously, nominal certificates have been used for reward shaping \cite{Cheng2019}, \cite{Westenbroek2021}, \cite{Westenbroek2022}. Previous work has also studied explicitly modifying the RL algorithm with barrier critics that are jointly learned with a safe policy \cite{Zhao2023}, \cite{yang2023feasible}, \cite{pmlr-v164-liu22c}, \cite{dalal2018safe}, \cite{liu2022safe}. Where previous work focuses on learning safe policies, we instead focus on studying conditions when learned value functions can be interpreted as control barrier functions, enabling transfer to downstream tasks. A survey of other approaches for safe learning-based control is available at \cite{Brunke2022Safe}.

\textbf{Certificate learning.} Generally, there exists a wealth of literature on learning of neural certificates \cite{Abate2021, Richards2018, Manek2020, Gaby2021}. While a full review of certificate learning is outside the scope of this paper, we refer interested readers to \citet{Dawson2022} for a comprehensive survey. Learning methods for neural certificates typically rely on self-supervised learning, consider continuous systems, assume knowledge of dynamics, and \textit{control-affine} dynamics. Certificates for discrete-time systems were studied in \cite{Grizzle2001, Dai2020}. Recent work studied certificate learning for black-box systems through learned dynamics models \cite{Qin2022}.  Compared to the main body of work on certificate learning, our method is applicable to a much wider range of systems as it works with black-box dynamics, discrete-time systems, and does not need control-affineness. 

\textbf{Constrained RL.} Finally, we discuss our work in the context of the constrained RL literature. As discussed in the introduction, such methods aim to preserve safety by augmenting $M$ with safety constraints of the form $c_i(x, u) \leq 0$. Methods for learning to solve cMDPs have been widely studied, such as Lagrangian methods \cite{Tessler2018, Stooke2020} and Lyapunov-based methods \cite{Chow2018}. More recent work considers building a trust region of policies \cite{Achiam2017, Zanger2021}, projecting to a safer policy \cite{Yang2020accelerating}, and using conservative policy updates \cite{Bharadhwaj2021}. Within this context, our results show that learned Q-functions can be directly used in a constraint of the form $c(x,u) = Q(x,u) - R \leq 0$ in order to guarantee safety. Hence, our method is orthogonal to and compatible with all of the constrained RL methods discussed above.   

\section{Limitations and Future Work}\label{section:limitations}

\textbf{Safety violations during exploration}. Our method assumes no prior knowledge on dynamics. Hence, a barrier function trained \textit{tabula rasa} will likely need to encounter many safety violations in order to learn safe behaviour. This may be unsuitable for learning in real-world environments where safety must be preserved throughout exploration. An exciting direction for future work is to reduce safety violations during exploration by using nominal (and possibly imperfect) dynamics models to pre-train a CBF solution using self-supervised learning approaches \cite{Dawson2022}, and subsequently fine-tune using our RL-based method. 

\textbf{Soft safety guarantees}. Despite empirically correlating well with the capacity of barrier functions to constrain unsafe policies, our validity metric can only be interpreted as a statistical argument for safety, rather than a formal proof; indeed, provable guarantees are impossible so long as we assume completely black-box dynamics. By considering gray-box dynamics models instead, such as nominal models with unknown parameters, future work can explore methods that provide stronger guarantees such as rigorous verification through Lipschitz-continuity bounds \cite{Dawson2022}, formal verification through symbolic logic \cite{Xie2022}, or exhaustive verification \cite{Albarghouthi2021}. 

\textbf{Offline learning}. In our work, we primarily consider learning the CBF separately from a downstream task policy. In future work, we hope to consider learning the CBF jointly with a task-specific safe policy in an online fashion. \cite{dalal2018safe}, \cite{pmlr-v164-liu22c}. 

\section{Conclusion}


This work presents theoretical contributions that establish a connection between barrier functions and value functions and demonstrates the feasibiliy of learning of barrier functions through an RL approach. We explore and ablate critical implementation details for learning high-quality barrier functions using our method. We demonstrate that standard RL metrics fail to evaluate the capacity of learned barrier functions to act as safety constraints. To address this gap, we propose our own novel barrier metrics. 

The proposed approach is especially suitable for learning \textbf{perceptual CBFs}, where safety can be defined as a direct function of sensor inputs. In one case study, perceptual CBFs on LiDAR scans enabled safe obstacle avoidance in cluttered environments \cite{Dawson2022perception}. 
In contrast to self-supervised learning, which requires careful handling of sensor dynamics, reinforcement learning naturally scales to end-to-end robot control \cite{Levine2016}, making it a  promising alternative. 

The theoretical contributions of this work have broad applicability and can extend to any MDP $M$ with any choice of RL algorithm. This suggests that our method can be employed to learn barrier functions for safe control in diverse tasks. Future work will extend to tasks with different reward structures by defiining an auxiliary safety-preserving reward for the unsafe set $\Xunsafe$ and training an auxiliary value function as the CBF. This will enable joint learning of safety constraints and task-oriented behaviours. 

In summary, our work contributes to the development of general, scalable, and \textit{verifiable} control methods that can be applied to various tasks. By introducing novel barrier metrics and leveraging reinforcement learning techniques, we provide a useful framework for developing verifiable control systems, enabling safer and more reliable autonomous behaviors in real-world environments.


\newpage
\bibliography{references_daniel,references_manual}

\newpage
\appendix
\onecolumn

\section{Preliminaries}\label{appendix:preliminaries}
\subsection{Markov Decision Processes}
A Markov Decision Process $M$ can be defined as a tuple (MDPs) $M = (\X, \U, f, r, \gamma)$, where $\X$ is the state space, $\U$ the control space, $f: \X \times \U \to \X$ the (discrete-time) dynamics, $r: \X \times \U \to [r_{min}, r_{max}]$ the reward function, and $\gamma \in [0,1)$ the discount factor. A trajectory $\tau$ is a sequence $\set{(x_t, u_t, r_t)}_{t \in \mathbb{N}}$ satisfying $x_{t+1} = f(x_t, u_t)$ and $r_t = r(x_t, u_t)$. A policy $\pi: \X \to \U$ induces associated \textit{closed-loop} dynamics $f_\pi(x) = f(x, \pi(x))$. 

\subsection{Reinforcement Learning}

Reinforcement learning (RL) is a broad family of algorithms designed to solve MDPs. Given a policy $\pi: \X \to \U$, it is common to define the Q-value function $Q_\pi$ and state value function $V_\pi$. 
\begin{align*}\label{def:q_value}
    & Q_\pi(x, u) = r(x, u) + \Big[\sum_{t = 1}^\infty \gamma^t r(x_t, \pi(x_t)) \Big] \\
    & V_\pi(x)  = \sup_u \set{Q_\pi(x,u)}
\end{align*}

The optimal $Q^*, V^*$ (for the reward-maximizing policy $\pi^*$) satisfy the one-step Bellman equality: 
\begin{equation}\label{eqn:Q_value_opt}
Q^*(x, u) = r(x, u) + \gamma  V^*(f(x,u))
\end{equation}
\section{Training History}\label{appendix:figures}
\begin{figure*}[h]
    \centering
    \includegraphics[width=0.48\linewidth]{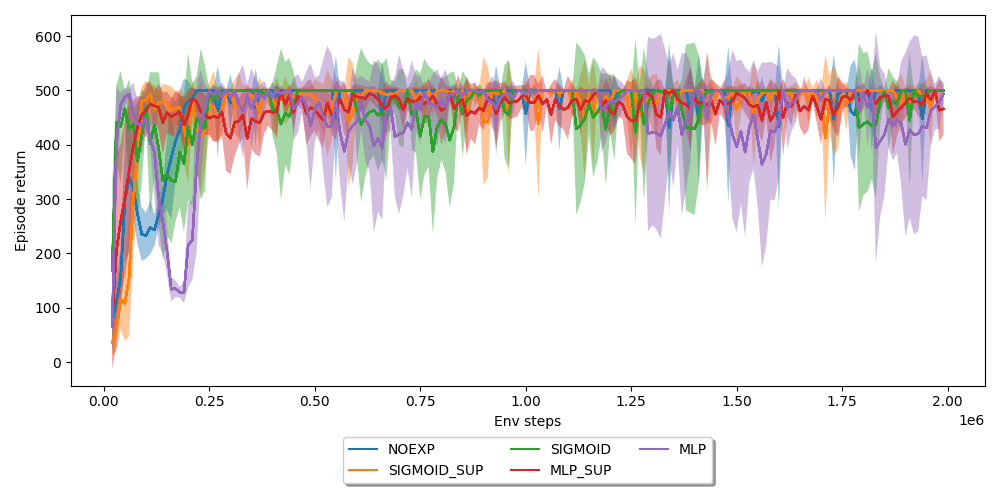}
    \includegraphics[width=0.48\linewidth]{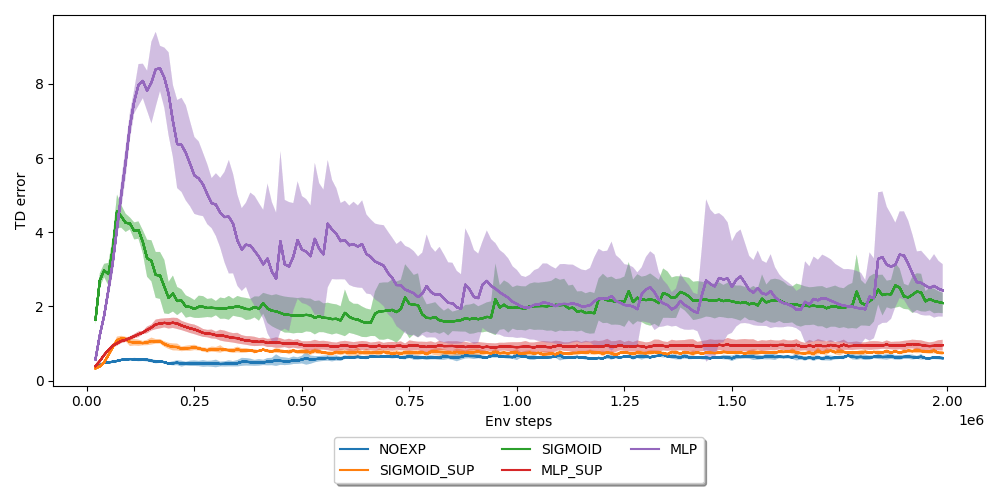}
    \caption{Left: Mean episode return over $10$ rollouts. In all cases, the $Q$-greedy policy achieves the maximum return of $500$. Right: Mean TD error across $n=10,000$ points sampled uniformly from the state space. The architectures with bounded parametrization achieve a lower mean TD error.} 
    \label{fig:train_hist_cbf}
\end{figure*}

\begin{figure*}[h]
    \centering
    \includegraphics[width=.48\linewidth]{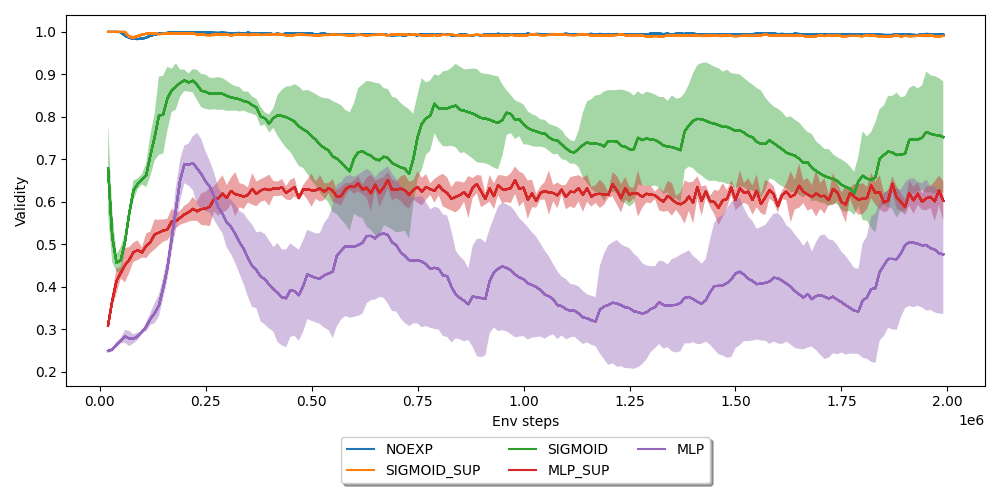}
    \includegraphics[width=.48\linewidth]{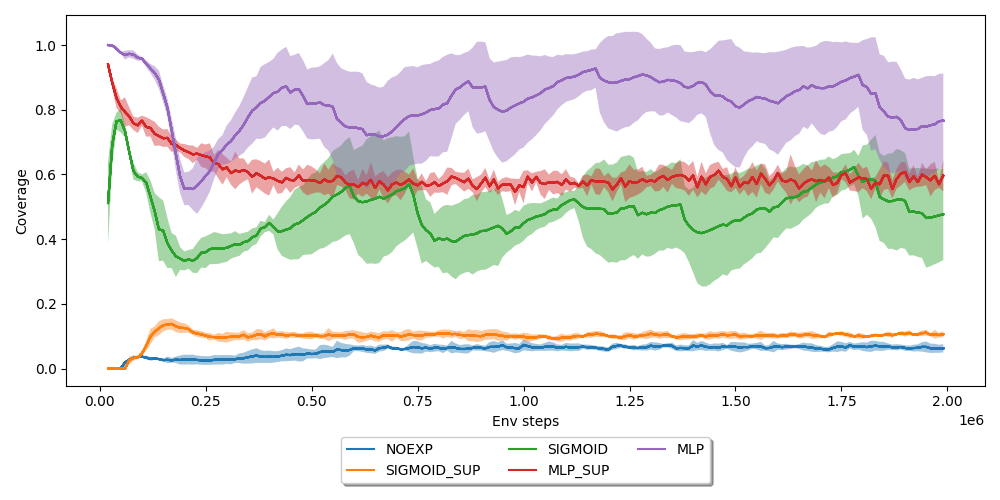}
    \caption{Training history of $m_{\mathrm{valid}}$ (top) and $m_{\mathrm{cov}}$ (bottom). The bounded and supervised value networks achieved the highest validity of approximately $100\%$. Enabling exploration increased coverage.}
    \label{fig:train_hist_validity_coverage}
\end{figure*}

\section{CartPole Schematic}\label{appendix:cartpole}
We provide a schematic of the CartPole environment
\begin{figure}[h!]
    \centering
    \includegraphics[width=0.5\linewidth]{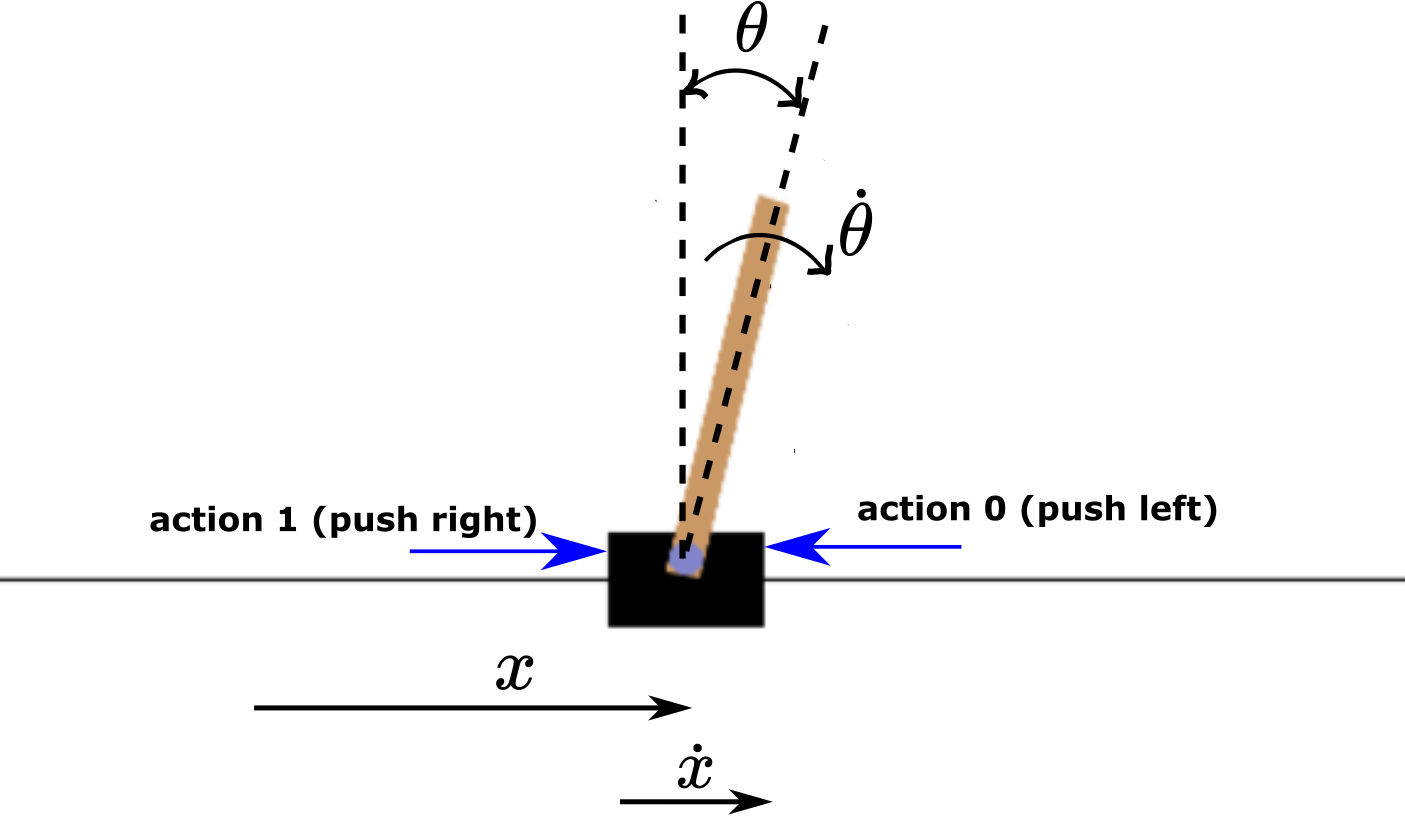}
    \caption{Schematic of the CartPole environment. The state space is parametrized as $(x, \dot{x}, \theta, \dot{\theta})$, where $x$ is the cart position and $\theta$ is the pole angle. We use the default termination condition as the unsafe set $\Xunsafe = \set{|x| \geq 2.4 \text{ or } |\theta| \geq 12^\circ}$. }
    \label{fig:cartpole}
\end{figure}
\section{Proofs}\label{appendix:proofs}
\textbf{Proof of Lemma \ref{thm:h_properties}}. 
\begin{proof}
Since $\alpha \leq 1$, if $h(x) \geq 0$ then $\sup_u \set{h(f(x,u))} \geq 0$. Thus, we can define $\pi(x) = \mathrm{argmax} \, h(f(x,u))$ and it is easy to see that $x \in \Xsafe \implies f(x, \pi(x)) \in \Xsafe$. 
\end{proof}
\textbf{Proof of Lemma \ref{thm:h_transform}}. 
\begin{proof}
We prove that $\tilde{h} = w \circ h$ satisfies both conditions discussed in (\ref{def:cbf}) to be a valid CBF. First, note:
\begin{align*}\label{proof:h_transform_i}
x \in \Xunsafe & \implies h(x) < 0 & \text{ by (\ref{def:cbf})(i)}\\ 
& \iff w(h(x)) < 0 & \text{ by (\ref{eqn:h_transform})(iii)}
\end{align*}
Hence $\tilde{h}$ satisfies (\ref{def:cbf})(i). Note that $w(h(x) \geq 0) \iff h(x) \geq 0$. Now, for $x: \tilde{h}(x) \geq 0$:
\begin{align*}
\sup_u \set{w(h(f(x,u))) - w(h(x))} & \geq C \sup_u \set{h(f(x,u)) - h(x)} & \text{ by (\ref{eqn:h_transform})(ii)} \\ 
& \geq -C \alpha h(x) &  \text{ by (\ref{def:cbf})(ii)} \\
& \geq - \alpha w(h(x)) &  \text{ by (\ref{eqn:h_transform})(i)} 
\end{align*}
Hence $\tilde{h}$ satisfies (\ref{def:cbf})(ii).  
\end{proof}

\textbf{Proof of Theorem \ref{thm:v_as_cbf}} 
\begin{proof}
    We consider the two conditions presented in (\ref{def:cbf}) that CBFs must satisfy. From (\ref{eqn:v_partitioned}), it is clear that (\ref{def:cbf})(i) is satisfied. Similarly, we note that $h(x) \geq 0$ implies $x$ is indefinitely safe; then by definition there exists a control such that $h(f(x,u)) = \frac{1}{1-\gamma} - R \geq (1-\alpha)(\frac{1}{1-\gamma} - R) = (1-\alpha)h(x)$. This proves condition (\ref{def:cbf})(ii).
\end{proof}

\textbf{Proof of Theorem \ref{thm:v_learned_as_cbf}} 
\begin{proof}
We prove that $h = V - R$ satisfies both conditions discussed in (\ref{def:cbf}) to be a valid CBF. First, let $x \in \Xunsafe$; then $V(x) \leq V^*(x) + \epsilon = \frac{1 - \gamma ^H}{1 - \gamma} + \epsilon < R$. Hence $h(x) < 0$ and condition (\ref{def:cbf})(i) is satisfied. 

Now, let $h(x) \geq 0$. Then $x \in \Xsafe$, thus $\sup_u h(f(x,u)) \geq V^*(x) - \epsilon - R = \frac{1}{1-\gamma} - \epsilon - R$. Similarly, we have $h(x) \leq V^*(x) + \epsilon - R = \frac{1}{1-\gamma} + \epsilon - R$. Then, to satisfy condition (\ref{def:cbf})(ii), it suffices that: 
\begin{align*}
& \frac{1}{1-\gamma} - \epsilon - R \geq (1-\alpha)(\frac{1}{1-\gamma} + \epsilon - R) \\
\implies & \alpha \geq \frac{2\epsilon}{\frac{1}{1-\gamma} + \epsilon - R}
\end{align*}
Note that this can be satisfied because $R < \frac{1}{1-\gamma} - \epsilon$; hence the R.H.S is strictly smaller than $1$. Hence condition (\ref{def:cbf})(ii) is satisfied under assumptions (a)-(e), which completes the proof. 
\end{proof}

\textbf{Proof of Remark \ref{remark:w_cbf_preserving}}
We first note that $g^{-1} = (1-\gamma)x$ is CBF preserving with $C = (1-\gamma)$. Conditions (\ref{thm:h_transform})(i), (ii), (iii) are all trivially verifiable. 

Next, we show that $\tilde{\sigma}^{-1}$ is CBF-preserving with $C = 1$, where $\tilde{\sigma}(x) = \sigma(x) - 0.5$. First, note that $x \geq 0 \iff \sigma(x) \geq 0.5$ and hence (i) is satisfied by substituting $x \to \tilde{\sigma}^{-1}(x)$. Next, note the identity $\sigma(x) \leq x + 0.5$; this implies $\tilde{\sigma(x)} \leq x$. Again, by substituting $x \to \tilde{\sigma}^{-1}(x)$, we observe that condition (ii) is satisfied. Lastly, note that $\sigma$ is Lipschitz-continuous with $L = 1$. This implies that $\sigma(x) - \sigma(y) \leq x-y$ for $x > y$; hence $\tilde\sigma(x) - \tilde\sigma(y) \leq x-y$. By substituting $x \to \tilde{\sigma}^{-1}(x), y \to \tilde{\sigma}^{-1}(y)$ we see that condition (iii) is satisfied. 

Lastly, since both $g^{-1}, \tilde{\sigma}^{-1}$ are CBF-preserving, if $h$ is a CBF then $g^{-1} \circ h$ is also a CBF, and then $\tilde{\sigma}^{-1} \circ g^{-1} \circ h$ is also a CBF. Hence $w$ is CBF-preserving as claimed.

\end{document}